\documentclass[10pt]{article}

\usepackage{graphicx}
\def\ind{{\mathchoice {\rm 1\mskip-4mu l} {\rm 1\mskip-4mu l}
{\rm 1\mskip-4.5mu l} {\rm 1\mskip-5mu l}}}
\usepackage{amsmath,amssymb}

\newtheorem{thm}{Theorem}
\newtheorem{example}{Example}
\newtheorem{appli}{Applications}
\newtheorem{defi}{Definition}
\newtheorem{prop}{Proposition}
\newtheorem{rem}{Remark}

\usepackage{algorithm,algorithmic}

\begin{document}

\title{Mean-field learning: a survey }

\author{Hamidou Tembine, Raul Tempone, Pedro Vilanova
\thanks{We are  grateful to the audience at Valuetools 2012 tutorial and many seminar participants for their valuable comments and suggestions on the preliminary version of the present work.}  \thanks{ The research of the first author has been supported by the ERC Starting Grant  305123 MORE (Advanced Mathematical Tools for Complex Network Engineering).}
\thanks{H. Tembine is with Ecole Superieure d'Electricite, Supelec, France
              E-mail: tembine@ieee.org,
           R. Tempone and P. Vilanova are with
              KAUST, Department of Applied Mathematics and Computational Science, Thuwal, Kingdom of Saudi Arabia}
              }

\maketitle


\begin{abstract}
In this paper we study iterative procedures for stationary equilibria in games with large number of players.
Most of  learning algorithms for games with  continuous action spaces are limited to strict contraction best reply maps in which the Banach-Picard iteration converges with geometrical convergence rate. When the best reply map is not a contraction, Ishikawa-based learning is proposed. The algorithm is shown to behave well for Lipschitz continuous and pseudo-contractive maps. However, the convergence rate is still unsatisfactory. Several acceleration techniques are presented. We explain how cognitive users can improve the convergence rate based only on few number of measurements. The  methodology provides nice properties in mean field games where the payoff function depends only on own-action and the mean of the mean-field. A learning framework that exploits the structure of such games, called, mean-field learning, is proposed. The proposed mean-field learning framework is suitable not only for games but also for non-convex global  optimization problems.
 Then, we introduce mean-field learning without feedback and examine the convergence to equilibria in beauty contest games, which have interesting applications in financial markets. Finally, we provide a fully distributed mean-field learning  and its speedup versions for satisfactory solution in wireless networks. We illustrate the convergence rate improvement with  numerical examples.
\end{abstract}

\section{Introduction}
Recently there has been renewed interest in large-scale games in several research disciplines, with its uses in financial markets, biology, power grid and cloud networking.  Classical work provides rich mathematical foundations and equilibrium concepts, but relatively little in the way of learning, computational and representational insights that would allow game theory to scale up to large-scale systems.
The literature on learning in games with discrete action space  is huge (see \cite{sergio,sorin,crc} and the references therein). However, only few results are available for continuous action space. In this paper, we explain how the  rapidly emerging field of mean-field games can address such behavioral, learning and algorithmic issues.

We propose both model-based (but still with less information) and non-model-based learning schemes for games with continuous action space and large number of players. Each player will update her learning strategies based on an aggregative term~\cite{aggregative}, which is the sum of  action of the other players. Each player will be influenced by the aggregate, and the mean field behavior is formed from the contributions of each player.
In the model-based mean-field learning scheme,  the mean action will be read by the players at each time slot, and each player will respond to the aggregative term locally. This simplifies drastically the dimensionally of the best-response system in the asymptotic case.

We distinguish different types of learning schemes depending on the information requirement:


 (i) Partially distributed strategic learning~\cite{crc}, where each player knows her own-payoff function, has some computational capabilities and observes the actions of the other players at the previous step.
    Examples of  such learning schemes include best response algorithms, fictitious play, and logit algorithms.

(ii) Fully distributed  strategic learning:
 In many dynamic interactions, one would like to have a learning and adaptive procedure which does not require
any information about the other players  actions or payoffs and less memory as possible (small number of parameters in term of past
own-actions and past own-payoffs). Fully distributed learning algorithms
  are only based on numerical measurements of signals or payoffs. The mathematical structure of own-payoff functions are not assumed to be known by the player. Hence, gradient-like ascent and best-reply algorithms cannot be used directly. The observations of private signals/measurements are not explicit in the actions of the other players.
 Based on numerical measurement of realized own-payoff, each player employs a certain learning pattern in order to learn the expected payoff function (payoff-learning) as well as the associated optimal strategies
    (model-free strategy-learning).  This type of learning algorithm is referred as Combined fully DIstributed PAyoff and Strategy learning (CODIPAS, \cite{bush,crc2,jsac}). These algorithms are simple but they play an important role in terms of applications since they are based on experiments and real data. In the continuous action space case, the gradient of own-payoff is not observed and hence it needs to be estimated or learned if one wants to use a gradient-like ascent method. However, estimating an accurate gradient based only on a sequence of payoff measurements is not a trivial task.

 (iii) No-feedback learning where the players do not observe any numerical payoff measurement. These schemes are based only on estimations and offline adjustment. However, conjectures and hierarchical reasoning could be used in order to get consistent reactions.

  In all the above three categories of  learning algorithms, the combination of the learning patterns of all the players form a multidimensional interactive system. 
The question we address is whether it is possible to exploit the structure of the payoff functions in large-scale regime to reduce the complexity of the above learning algorithms (partially or fully distributed).

The answer to this question is positive for some aggregative games. 
We examine three  classes of mean-field learning frameworks:
\begin{itemize}
\item {\it Partially distributed mean-field learning} where each player knows  her own-payoff function, has some computational capabilities and observes the mean field at the previous step.
    Examples of  such learning algorithms  include best  response to mean field and Boltzmann-Gibbs mean-field response.

\item {\it Fully distributed mean-field learning} schemes can be used in situations where each generic player in the large population is able to observe/measure a numerical value of her own-payoff (that could be noisy). These schemes are derivative-free and model-free. They can be applied in both mean-field games and mean-field global optimization.

\item {\it No-feedback mean-field learning}: There are some situations where it is difficult to feedback any information to the population of players and local  measurement of own-payoff is not available. Then, the above two classes of learning algorithms that are based on feedbacks to the players are inappropriate. In that case, a learning scheme without feedback can be employed if the payoff functions are common knowledge.
\end{itemize}

\subsection{Overview: learning for games with continuous action space}
We briefly overview fully distributed learning for games with continuous action space.
One of the first fully distributed learning algorithms is the so-called reinforcement learning. While there are promising results in Markov decision processes with few number of states and actions, majority of reinforcement  Q-learning, adaptive heuristic critic, and regret minimizing multi-arm bandit algorithms meet several difficulties in continuous action space.
The difficulty in extending such learning algorithms to multi-player games is that with a
balance has to be maintained between exploiting the information gained during learning, and
exploring the set of actions (with is a continuum) to gain more information. Instead of updating a finite dimensional probability vector, one has to adjust a probability density function in infinite dimensional space.
The authors in \cite{tref1} has proposed reinforcement learning algorithms for games with continuous and compact action space applied to vehicle suspension control. The convergence analysis is not conducted in \cite{tref1} .

In \cite{tref2} the authors studied continuous action reinforcement learning automata and applied to  adaptive digital filters. The authors claimed convergence of their algorithm  via computer simulations.

Recently, \cite{tref3} observed a poor performance and selection of basis functions that are used in \cite{tref1,tref2}  to approximate the infinite dimensional space. It is conjectured that the convergence time (when it converges) is very high even for quadratic cost function.

In order to reduce the dimensionality, the basic idea in these continuous action space reinforcement learning studies has been to use a normal distribution at each time slot and updates the mean and standard deviation based on the  payoff measurement.

Another approach to continuous action space learning is the regret minimizing procedure that allows to be close  to the Hannan set. Such a procedure have been widely studied for discrete action space and has several interesting computational properties: the set of correlated equilibria is convex and there is polynomial time algorithm. The extension to continuous and compact action space has been conducted in \cite{tref4}. It is shown that the empirical
frequencies of play converges to the set of correlated equilibria.
However, most of these convergence results are not for a point but a set and the convergence time is not provided in \cite{tref4}. Another important point is that the convergence of the frequency of play does not imply the convergence of actions or strategies. All the above references consider finite number of players.

In this work we are interested on  learning in games with {\it large number of players} and {\it continuous action space}. The framework presented here differs from classical machine learning for large-scale systems.
 The main difference is the strategic behavior of the players who make decisions in a distributed and autonomous manner. This creates an interdependency between the decisions through the mean of the mean-field.

\subsection{Contribution}
 Our contributions can be summarized as follows. First, we introduce a learning framework for games with large population of players, called mean-field learning. Considering payoff functions that depend on own-input and an aggregative term, we show that mean-field learning simplifies drastically the analysis of large-scale interactive learning systems.
In the single class case, it reduces to the analysis of one iterative process instead of an infinite collection of learning processes. Second, we study both asymptotic~\cite{mfg} and non-asymptotic properties  of the resulting learning framework. Stability, error bounds and acceleration techniques are proposed for model-based mean-field learning as well as for derivative-free mean-field learning. In particular, we show that the convergence time of $(o+1)-$order speedup learning is at most
$$
T_{\eta^*}=\frac{1}{\ln(o+1)}\ \ln\left[ \frac{\ln\left(\frac{1}{\eta^*c_2^{\frac{1}{o}}}\right)}{\ln\left(\frac{1}{\eta_0c_2^{\frac{1}{o}}}\right)} \right]
$$
where $\eta^*$ is the error target, $c_2$ is a positive value which does not depend on time and $\eta_0$ is the initial error gap.
Interestingly, the methodology extends to satisfactory solution (which is a situation where all the players are satisfied). We provide fully distributed mean-field learning schemes for satisfactory solution in specific games. Reverse Ishikawa and Steffensen speedup learning are proposed  to improve the convergence rate to satisfactory solution. Numerical examples of the basic learning techniques are illustrated and compared in guessing games and in quality-of-service (QoS) satisfaction problems in wireless networks.

\subsection{Structure}
The paper is organized as  follows. In the next section we present a generic mean-field game model. In Section \ref{thut} we present the  mean-field learning framework. In  Section \ref{exsec} we present a detailed example of beauty contest mean-field game. Speedup strategic learning  for satisfactory solution are presented in Section \ref{secsat}.
Finally, section \ref{cont} concludes the paper.

The proofs are given in Appendix.


\section{Mean field game model}
Consider $n\geq 2$ players. Each player takes her action in the convex set $\mathcal{A}\subseteq \mathbb{R}^d,\ d\geq 1.$ Denote the set of players by $\mathcal{N}=\{1,2,\ldots, n\}.$
In the standard formulation of a game, a player's payoff function depends on
opponents' individual actions. Yet, in many games, payoff functions depend
only on some aggregate of these, an example being the Cournot model where it
is the aggregate supply of opponents that matters rather than their individual
strategies. The main property of aggregative games is the structure of the
payoff functions. The payoff function of each player depends on its own action
and an aggregative term of the other actions. A generic payoff function in
aggregative game with additive aggregative term is given by $r_j:\ \mathcal{A}^2\longrightarrow \mathbb{R}$

\begin{eqnarray} \label{eqntype1}
\tilde{r}(a)=r_{j}\left( a_{j},\frac{1}{n-1}\sum_{j'\neq j}a_{j'}\right) ,
\end{eqnarray}
where the actions are real numbers and $\tilde{r}:\ \mathcal{A}^n\longrightarrow \mathbb{R}.$ In this context the key term of player $j$ is the structure of the function $r_j$, its own-action $a_j$ and the aggregative term
$\frac{1}{n}\sum_{j'=1}^n a_{j'}.$

The triplet $\mathcal{G}:=(\mathcal{N}, \mathcal{A}, (\tilde{r}_j)_{j\in\mathcal{N}})$ constitutes a one-shot game in strategic form.

\begin{appli} The type of aggregate-driven  reward function in (\ref{eqntype1}) has wide range of applications:

  (a) In economics and financial markets, the market price (of products, good, phones, laptops, etc) is influenced by the total demand and total supply,

 (b) In queueing theory, the task completion of a data center or a server is influenced by the mean of how much the other data centers/servers can serve.

 (c) In resource sharing problems, the utility/disutility of a player depends on the demand of the other players. Examples include cost sharing in coalitional system and  capacity and bandwidth sharing in cloud networking.

   (d) In wireless networks, the performance of a  wireless node is influenced by the interference created by the other  transmitters.

  (e) In congestion control, the delay of a network depends on the aggregate (total) flow and the congestion level of the links/routes.


\end{appli}


\begin{defi}
The action profile $(a_j)_{j\in\mathcal{N}}$ is a pure equilibrium of the game $\mathcal{G}$ if no player can improve her payoff by unilateral deviation i.e., for every player $ j\in\mathcal{N},$ one has
$$
\tilde{r}_j(a)\geq \tilde{r}_j(a'_j,a_{-j}),\ \forall a'_j\in\mathcal{A}
$$
\end{defi}

Before going for pure equilibrium seeking we first need  to ask if the problem is well-posed, i.e; the existence of a pure equilibrium. Below we provide a classical sufficiency condition for existence of a pure equilibrium in continuous-kernel  aggregative games.

The following results hold:
\begin{itemize}\item compactness: If $\mathcal{A}$ is a non empty, compact, convex subset of $\mathbb{R}^d$, and each $r_j$ is (lower semi-) continuous in $\mathcal{A}^2$ and quasi-concave with the respect to the first variable
then $\mathcal{G}$ possesses at least one pure strategy Nash equilibrium.
\item If $\mathcal{A}$ is  non-compact, we require additional coercivity assumption: $\ \tilde r_j(a_j, a_{-j})\longrightarrow -\infty$ as $\parallel a_j \parallel\longrightarrow +\infty.$
\end{itemize}

For discontinuous payoff function $r_j$ we refer to the recent development of existence of pure equilibria.
 A very active area is the full characterization for the existence of pure strategy
Nash equilibrium in games with general topological strategy spaces that may be discrete, continuum or non-convex and payoff functions that may be discontinuous or do not have any form of
quasi-convexity. For more details, see the literature review in Tian (2009, \cite{tian09}).

Below we examine only the cases in which the game admits at least one equilibrium and  the best-response is uniquely given by the mapping $f$ (which is not necessarily continuous).
We present  aggregate-based learning algorithms. Generically, a partially distributed mean-field-based learning scheme can be written as
    $
    a_{j,t+1}=F_j(a_{j,t},\ldots,a_{j,0}, \bar m_{n,t}, \ldots, \bar m_{n,0})
    $
    where $F$ is specified from the game model and  $\bar m_{n,t}$ is a mean action of the players at time $t.$ In this paper we examine only schemes with one-step memory   in the form $a_{j,t+1}=F_j(a_{j,t},\bar m_{n,t})$

 \subsection{Aggregate-based best-response}
 Let $\tilde{m}_{j,n}=\frac{1}{n-1}\sum_{j'\neq j}a_{j'}$ be the mean of actions of the others.
 We assume that $\arg\max_{a'_j}r_j(a'_j,\tilde{m}_{j,n})$ has a unique element which we denote by $\bar{f}_j(a_{-j}).$ Then, $\bar f(a)=(\bar f_j(a_{-j}))_{j}$ is the best-response map. We aim to find a fixed-point of such a map in $\mathcal{A}^n$ which is in an Euclidean space with dimension $nd.$ Exploiting the aggregative structure of the game and the convexity of the set $\mathcal{A},$ the  domain of the function $\bar f_j$ is reduced to the set $\mathcal{A}$ by the following relation $\bar f_j(a_{-j})=f_j(\tilde{m}_{j,n})$ where $\tilde{m}_{j,n}\in \mathcal{A}$ by convexity of the domain $\mathcal{A}.$
 The simultaneous best-response algorithm, called mean-field response, is given in Algorithm \ref{algobr1} which requires that player $j$ observes the common term $\bar{m}_{n,t-1}=\frac{1}{n}\sum_{j'=1}^n a_{j',t-1}$ at the previous step and 
 computes $$ \tilde{m}_{j,n,t-1}=\frac{1}{n-1}\left( n\bar{m}_{n,t-1}-a_{j,t-1}\right).$$

 \begin{algorithm}[!htp]
\caption{\textbf{:} Mean-field response}
\begin{algorithmic}[1]\label{algobr1}
\STATE \textbf{Initialization :}\\
for each user $j\in \mathcal{N}$ initialize ${a}_{j,0},$\\ 

\STATE \textbf{Learning pattern :}\\
for each time slot $t$\\
for each user $j\in \mathcal{N}$ do\\ 
Observe the aggregate $\bar{m}_{n,t}$\\
$a_{j,t+1}=f_j(\tilde{m}_{j,n,t})=f_j\left( \frac{1}{n-1}( n \bar{m}_{n,t}- a_{j,t})\right)$
\end{algorithmic}
\end{algorithm}

\subsection{Banach-Picard learning algorithm}
One of the first basic iterative procedures for finding fixed-point of the continuous map $f$ over complete metric space $\mathcal{A}$ is the Banach-Picard iterate. 
The algorithm consists to start at some point $a_0\in \mathcal{A},$ and take the compositions $f(a_0), f(f(a)),\ldots, $ that is $a_{t+1}=f(a_t)$, where $f:\ \mathcal{A}\longrightarrow \mathcal{A}.$ This algorithm is known to be convergent for strict contraction map, i.e., if there exists a Lipschitz constant $0<L<1$ of the function $f$ then the iterates converge (with geometric convergence rate $L$) to the unique fixed-point of $f$ in $\mathcal{A}.$
However, in many applications of interest the function $f$ may not be a contraction. See Example \ref{exampleont1} below.


\begin{thm}[\cite{za2}] \label{thmrs}
Let $(\mathcal{A},d)$ be a complete metric space, and $f : \mathcal{A} \longrightarrow \mathcal{A}$ a map for which
there exist real numbers $\alpha_1,\alpha_2$ and $\alpha_3$ satisfying $0 < \alpha_1< 1, 0 < \alpha_2,\alpha_3 < 1/2$ such that for each pair
$a_1, a_2$ in $\mathcal{A}$, at least one of the following conditions is true:
\begin{eqnarray}
(C0) & d(f(a_1),f(a_2)) \leq  \alpha_1d(a_1, a_2); \nonumber\\
(C1) & d(f(a_1),f(a_2)) \leq  \alpha_2[d(a_1,f(a_1))+d(a_2,f(a_2))];\nonumber\\ \nonumber
(C2) & d(f(a_1),f(a_2)) \leq  \alpha_3[d(a_1,f(a_2))+d(a_2,f(a_1))].
\end{eqnarray}
Then, the Banach-Picard algorithm converges to a fixed-point of $f$ for any initial point $a_0\in\mathcal{A}.$ Moreover, $f$ has a unique fixed-point.
\end{thm}

Note that the first part of Theorem \ref{thmrs} C0 has many applications in solving nonlinear equations, but suffers
from one drawback - the contractive condition (C0) forces $f$ to be continuous on $\mathcal{A}.$ However, the second condition (C1) does not require continuity of the best response function $f.$
The condition (C1) have been studied by Kannan (1968, \cite{kan68}) and (C2) by Chatterjea (1972, \cite{chat72}).

\begin{example}[Resource sharing game] \label{exampleont1}
Consider $n$ players in a  network with a resource capacity $c_n>0.$ Each player has a certain demand $a_j\geq 0$ which corresponds to her action. The action space is $\mathbb{R}_{+}.$ Denote by $p_n>0$ the unit price for each resource utilization. The payoff function is $$r_j(a)=c_n\frac{a_j}{\epsilon_n+\sum_{j=1}^n a_j}-p_n a_{j},$$ where $\epsilon_n$ is a positive parameter. Then, the (simultaneous) best-response algorithm is given by
$$
a_{t+1}=\left[\sqrt{\frac{c_n}{p_n}(\epsilon_n+(n-1)a_{t-1})}-(\epsilon_n+(n-1)a_{t-1})\right]_{+}
$$
and
$$f(a)=\left[\sqrt{\frac{c_n}{p_n}(\epsilon_n+(n-1)a)}-(\epsilon_n+(n-1)a)\right]_{+}.$$ A direct computation of the  derivative (at the interior) gives
$$
f'(a)=-(n-1)+\frac{c_n(n-1)}{p_n}\left(\frac{c_n}{p_n}(\epsilon_n+(n-1)a)\right)^{-1/2}.
$$
Clearly, $f$ is not a  contraction.
\end{example}
\begin{example}\label{exampleont12}
A non-convergent Banach-Picard iteration is obtained for $\mathcal{A}=[\frac{1}{4}, 4],$ and $f(a)=1/a.$ $f$ is $16-$Lipschitz in the domain $\mathcal{A}.$  Let start with $a_0\neq 1$ then $a_{2t}=a_0,\ a_{2t+1}=\frac{1}{a_0}\neq a_0.$ The two subsequences $a_{2t}$ and $a_{2t+1},\ t\geq 0$ have different limits. Hence,
the sequence $a_{t+1}=f(a_t)$ does not converge if the starting point is $a_0\neq 1.$
\end{example}
We explain below how to design a convergent sequence for the problem of Examples \ref{exampleont1} and \ref{exampleont12}.
A simple modification of Banach-Picard consists to introduce a learning rate $\lambda$ which takes the average between the previous action and the best response, $a_{t+1}=\lambda f(a_t)+(1-\lambda)a_t.$ Then, the procedure evolves slowly. The idea goes back at least to Mann (1953, \cite{mann}) and Krasnoselskij (1955, \cite{Kra55}).
For $\lambda$ equal to one, one gets the Banach-Picard algorithm.

If the function $f$ satisfies (Lipschitz and strongly pseudo-contractive map)
\begin{eqnarray}
(C3)\ \Vert a_1{-}a_2\Vert \leq \Vert a_1{-}a_2{+}s[a_1{-}f(a_1){-}k a_1{-}  (a_2{-}f(a_2){-}k a_2)]\Vert \nonumber
\end{eqnarray}
for any pair $(a_1, a_2)\in \mathcal{A}^2,$ where $s>0,k>0,$
and there is $\bar{\lambda}$ such that for any $\lambda$, $0<\lambda< \bar{\lambda},$ the learning algorithm  converges to a fixed-point.
However, for $\lambda$ closer to one, the algorithm may oscillates around a fixed-point.
In order to get a smaller $\lambda$ in the long-run, we can attempt to take a decreasing learning rate $\lambda_t\longrightarrow 0$ as $t$ grows. $\sum_{t=1}^{\infty}\lambda_t=+\infty.$

The algorithm (also referred to as Mann's algorithm \cite{mann}) reads
\begin{eqnarray}
a_{t+1}=\lambda_t f(a_t)+(1-\lambda_t)a_t\\
0<\lambda_t< 1,\
a_0\in \mathcal{A}.
\end{eqnarray}
\begin{algorithm}[!htp]
\caption{\textbf{:}  Mann-based mean-field response}
\begin{algorithmic}[1]\label{algobr2}
\STATE \textbf{Initialization :}\\
for each user $j\in \mathcal{N}$ initialize ${a}_{j,0},$\\ 
Define the sequence up to $T:\ \lambda_{j,t}$ for $\ t\in\{1,2,\ldots,T\}$\\ 
\STATE \textbf{Learning pattern :}\\
for each time slot $t$\\
for each user $j\in \mathcal{N}$ do:\\ 
Observe the aggregate $\bar{m}_{n,t}$\\
$a_{j,t+1}=\lambda_{j,t}f_j[\frac{1}{n-1}( n\bar{m}_{n,t-1}-a_{j,t-1})]+(1-\lambda_{j,t})a_{j,t}$
\end{algorithmic}
\end{algorithm}

If the function $f$ does not satisfy $C0,$ one can still get some convergence results due to Ishikawa \cite{ishi74,ishikawa}.

\begin{eqnarray}
a_{t+1}=\lambda_t f\left( \mu_t f(a_t)+(1-\mu_t)a_t
\right)+(1-\lambda_t)a_t\\
0<\lambda_t< 1,\
0\leq \mu_t \leq 1.\\
a_0\in \mathcal{A}.
\end{eqnarray}

Clearly, the same technique can be extended to a finite number of compositions of the mapping $f.$

\begin{algorithm}[!htp]
\caption{\textbf{:}  Ishikawa-based mean-field response}
\begin{algorithmic}[1]\label{algobr4isi}
\STATE \textbf{Initialization :}\\
for each user $j\in \mathcal{N}$ initialize ${a}_{j,0},$\\ 
Define the sequence up to $T:\ \lambda_{j,t}$ for $\ t\in\{1,2,\ldots,T\}$\\ 
\STATE \textbf{Learning pattern :}\\
for each time slot $t$\\
for each user $j\in \mathcal{N}$ do:\\ 
Observe the aggregate $\bar{m}_{n,t}$\\
$a_{j,t+1}=\lambda_{j,t}f_j(y_{j,t})+(1-\lambda_{j,t})a_{j,t}$\\
$y_{j,t}=\mu_{j,t}f_j\left(\frac{1}{n-1}( n\bar{m}_{n,t-1}-a_{j,t-1})\right)+(1-\mu_{j,t})a_{j,t}$
\end{algorithmic}
\end{algorithm}
\begin{figure}[htb]
\centering
  \includegraphics[width=0.7\columnwidth]{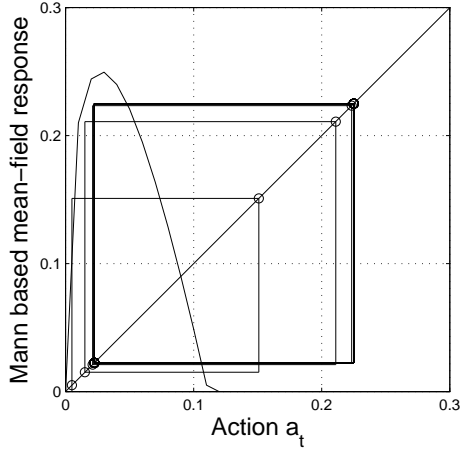}\\
  \caption{Mann-based mean-field response for example \ref{exampleont1} with big learning rate: presence of limit cycle.}\label{fig1cycle}
\end{figure}

\begin{figure}[htb]
\centering
  \includegraphics[width=0.7\columnwidth]{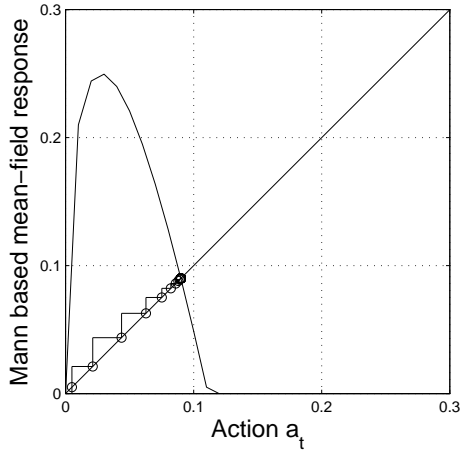}\\
  \caption{Ishikawa's based mean-field response for example \ref{exampleont1} with small learning rate.}\label{fig2nocycle}
\end{figure}

Figure \ref{fig1cycle} illustrates a cycling behavior of the mean-field response of example \ref{exampleont1}. The parameters are $\mu_{j,t}=0, \lambda_t=0.9, n=10, \epsilon_n=0, c_n=1, p_n=1, a_0=0.005.$ For $\lambda=0.1$ smaller than the one in Figure \ref{fig1cycle}, the cycle disappeared and the Ishikawa's based mean-field response scheme behaves well. See Figure \ref{fig2nocycle}.
\subsection{Faster algorithms: Banach-Picard vs Ishikawa}
For a class functions satisfying one of the conditions C0-C2, we know from Theorem \ref{thmrs} that there is a unique fixed-point of $f$ and the speed of convergence of the algorithm can be compared for different parameter $\mu_t.$
Suppose that $a_t$ and $b_t$ are two real convergent sequences with
limits $a^*$ and $b^*$ respectively. Then $\{a_t\}_t$ is said to converge faster than $\{b_t\}_t$ if
$\lim_t \frac{d(a_t,a^*)}{d(b_t,b^*)}=0
$

The authors in \cite{za,za2} showed that for a particular class of functions that satisfies one of the conditions C0, C1 or C2, the Banach-Picard algorithm is faster than the Ishikawa's algorithm with  $\mu_t=0$ is faster than the one with $\mu_t>0.$ However in general these algorithms are not comparable due to non-convergence. For example for
$f(a)=1/a, \ \mathcal{A}=[1/4, 4]$ there is a unique fixed point but Banach-Picard does not converge starting from $a_0\neq 1.$ However, the Ishikawa method converges to $1$ for $\lambda$ small enough.

\subsection{Reverse Ishikawa mean-field learning}
The reverse Ishikawa's mean-field learning consists to choose a learning rate (bigger than one) that converges to one.
\begin{eqnarray}
a_{t+1}=\lambda_t f(a_t)+(1-\lambda_t)a_t\\
1<\lambda_t< 2,\ \lim_t \lambda_t=1,\
a_0\in \mathcal{A}.
\end{eqnarray}
Example of $\lambda_t$ could be $1+\frac{1}{1.5^t}.$ When convergent, this scheme has the advantage of being faster than the fixed-point iteration.

\subsection{Non-asymptotic properties}
We focus on the non-asymptotic properties of the Ishikawa-based learning algorithms. Non-asymptotic strategic learning is very important in engineering applications.
 Traditional results on the fundamental limits of data compression and data transmission through noisy channels apply to the asymptotic regime as codelength, blocklength goes to infinity.  However, these asymptotic results are not usable if the window size and horizon are bounded. Therefore, it is interesting to look at the non-asymptotic regime of learning algorithms.
We provide generic rate of convergence of some class of best-response functions.

\subsubsection{Strict contraction}
For strict contraction mapping $f$ with constant $\alpha_1=L<1$, one has the following estimates:
$$
d(a_t,a^*)\leq \frac{\alpha_1^t}{1-\alpha_1}d(a_0,a^*),\ t>0.
$$
The advantage of this inequality is that it provides a error gap at any time which is a non-asymptotic learning result.

\subsubsection{Nonexpansive best-response function}
The next result provides the convergence rate of the asymptotic regularity for nonexpansive maps i.e., the class of map with Lipschitz constant $\alpha_1=1.$ Denote by
$diameter(\mathcal{A})$ the diameter of the set $\mathcal{A}.$
\begin{thm}[\cite{baillon}] \label{thmbaillon}
Let $f: \mathcal{A}\longrightarrow \mathcal{A}$ is a nonexpansive self-mapping of
a bounded convex subset $\mathcal{A}$ of a normed linear space, normalized so $diameter(\mathcal{A}) <+\infty$ with non-empty fixed-point set $fix(f);$
the Ishikawa algorithm with $\mu_t=0$ and $\lambda_t=\lambda\in (0,1),$ is a proper convex combination then
$$
d(a_t,f(a_t))\longrightarrow 0.
$$

Moreover
$$d(a_t,f(a_t))\leq \frac{diameter(\mathcal{A})}{\sqrt{\pi \sum_{t'=1}^t\lambda_{t'}(1-\lambda_{t'})}}.$$

If $\mathcal{A}$ is unbounded, the following estimate holds:
$$d(a_t,f(a_t))\leq 2\frac{d(a_0, fix(f))}{\sqrt{\pi \sum_{t'=1}^t\lambda_{t'}(1-\lambda_{t'})}}.$$
\end{thm}
We observe that when $\sum_{t'=1}^t\lambda_{t'}(1-\lambda_{t'})\longrightarrow +\infty$ one gets the so-called asymptotic regularity of the sequence generated by the Ishikawa algorithm.



\subsubsection{Strongly pseudocontractive}
\begin{thm}[\cite{berinde}] \label{pththm3}
Let $f$ Lipschitz with constant $L$ and strongly pseudocontractive with constant $k$ such that $fix(f)$ is non-empty. Then the Ishikawa algorithm with $\lambda_t=\lambda\in (0,\bar \lambda),
\bar\lambda=\frac{k}{(L+1)(L+2-k)},\ \mu_t=0$  converges strongly to the (unique) fixed point of $fix(f).$ The convergence rate is geometric and is given by
$$
d( a_t, fix(f))\leq d(a_0,fix(f)) \rho(\lambda)^t
$$
where $$\rho(\lambda)=\frac{1 + (1-k)\lambda + (L + 1)(L + 2 -k)\lambda^2}{1+\lambda},$$ which is minimized for $\lambda^*=-1+\sqrt{1+\bar\lambda}.$
\end{thm}

\subsection{Asymptotic pseudo-trajectories}
Using classical approximation the asymptotic pseudo trajectories of Ishikawa's algorithm can be studied using ordinary differential equations (ODEs). We assume that the function $f$ has a unique integral curve, in order to guarantee the uniqueness of the ODE (given a starting point).

\subsubsection{$\mu_t=0$}

The ODE is given by

$$\dot{a}_j=\frac{d}{dt}a_j=f_j(\tilde{m}_{j,n})-a_{j},$$ which is the aggregate-response dynamics in continuous time.

If $f_j$ denotes a best response of $\tilde{m}_{j,n}$ then the steady state of the aggregate-response dynamics are Nash equilibrium of the aggregative game.
If the dynamics has a Lyapunov function then one gets a global convergence to Nash equilibria. These  games are called {\it Lyapunov games}.


Now we turn to the convergence time for a given error/precision.
We define the convergence time $T_{\eta}$ of $a_t$ within a $\eta-$neighborhood as the first time the trajectory of pure strategy reaches a neighborhood of range $\eta$ to the set of fixed-points.
$$
T_{\eta}=\inf\{ t\geq 0,\ |  \ d(a_t,fix(f))\leq \eta \}.
$$

\subsubsection{How to accelerate the convergence time of the ODE?}
Consider the ordinary differential equations (ODEs) that captures the trajectories of the learning pattern in transient phase:  $\dot{a}_t=f(a_t)$ and $\dot{b}_t=\lambda_t f(b_t).$ Assume that the two ODEs start at the same point $a_0=b_0.$ What can we say about the trajectories of $a$ and $b?$

We explicitly give the convergence  time of $b$ in function of  that of $a.$
\begin{prop}\label{abconv}
The explicit solution is given by
$$b_t=a_{\int_0^t \lambda_s\ ds}.$$
In particular,  if the trajectory of $a$ reaches a target set $O$ for at most  $T_a$ time units then the trajectory of $b$ reaches the same set for at most $T_b=g^{-1}(T_a)$ where $g:\ t\longrightarrow \int_0^t \lambda_s\ ds.$

If $\lambda_s=\lambda,\ $ then $T_b=\frac{T_a}{\lambda}.$

If $\lambda_s=e^{s}, $ then $T_b{=}\ln(T_a+1),$ i.e., $\frac{T_b}{T_a}$ goes to zero and $b$ is faster than $a.$
\end{prop}

\section{Mean-field learning} \label{thut}
We now consider a continuum of players. The mean-field is the action distribution $m.$ Its mean is in the set $\mathcal{A}$ and is denoted by $\bar{m}.$
Then the limiting payoff for single class writes $r(a,\bar{m}).$ All the functions $f_j$ above are reduced a single function $f(\bar{m}),$ i.e., each player responds to the mean of the mean-field.
Therefore the Banach-Picard (mean-field response) dynamics becomes
$$
a_{t+1} =\bar{f}(a_t, \bar{m}_t)
$$
which can be reduced to
\begin{eqnarray}
a_{t+1} =\bar{f}(a_t, \bar{m}_t)=\bar{m}_{t+1}=\bar{f}(\bar{m}_t, \bar{m}_t)=:f(\bar{m}_t)
\end{eqnarray}
by indistinguishability property.

Thus, Banach-Picard-based mean-field response learning is given by
\begin{eqnarray}\label{mfbr}
\bar{m}_{t+1}=f(\bar{m}_t),
\end{eqnarray}
Note that in  Equation (\ref{mfbr})  only the starting point $\bar{m}_0$ is required if the player knows the structure of $f.$ This means that  it is  not needed to feedback the mean of the mean-field at each step.
\begin{prop}\label{proth1}
Consider a mean field game $r(a,\bar{m})$ such the mean field response function $f$ is a strict contraction mapping over $\mathcal{A}$ (non-empty, convex subset of $\mathbb{R}^d$) with constant $\alpha_1$.
Then, the mean-field response learning (\ref{mfbr})
finds an approximated fixed-point within a $\eta-$neighborhood in at most $T_{\eta}=1+\lfloor \max(0,T)\rfloor$  number of iterations where $T=
\frac{\ln [ \frac{d(\bar{m}_0, fix(f))}{\eta (1-\alpha_1)}]}{\ln \frac{1}{\alpha_1}}.$
\end{prop}

The Ishikawa-based mean field response is
\begin{eqnarray}\label{mfishi}
\bar{m}_{t+1}=\lambda_t f\left[ \mu_t f(\bar{m}_t)+(1-\mu_t)\bar{m}_t
\right]+(1-\lambda_t)\bar{m}_t\\
0<\lambda_t< 1\ , 0\leq \mu_t \leq 1\\
\bar{m}_0\in \mathcal{A}.
\end{eqnarray}

Based on  Theorem \ref{thmbaillon}, the next Proposition provides an upper bound of the convergence time in order of $O(\frac{1}{\eta^2}).$
\begin{prop} \label{th2time}
Consider a mean field game $r(a,\bar{m})$ such the mean field response function $f$ is a nonexpansive mapping over $\mathcal{A}$ (non-empty, convex subset of $\mathbb{R}^d$).
Then, the Ishikawa based mean-field learning (\ref{mfishi})
finds an approximated fixed-point within an $\eta-$neighborhood in at most $T_{\eta}=1+\lfloor \max(0,T)\rfloor$ number of iterations where $T=\frac{16 d(\bar{m}_0, fix(f))^2}{\eta^2\pi}$.
\end{prop}

Proposition \ref{proppseudo} provides a convergence time bound in order of $O( \ln (\frac{1}{\eta}))$ for Lipschitz with constant $L$ and strongly pseudocontractive mean-field response function $f.$
\begin{prop} \label{proppseudo}
Let $f$ Lipschitz with constant $L$ and strongly pseudocontractive with constant $k$ such that $fix(f)$ is non-empty. Then the Ishikawa algorithm with $\lambda_t=\lambda\in (0,\bar \lambda),
\bar\lambda=\frac{k}{(L+1)(L+2-k)},\ \mu_t=0$  converges strongly to the (unique) fixed point of $fix(f)$ with at most
$T_{\eta}=1+\lfloor \max(0,T)\rfloor$ number of iterations where $T=\frac{\ln [ \frac{d(\bar{m}_0, fix(f))}{\eta }]}{\ln \frac{1}{\rho(\lambda^*)}},$
 $\rho(\lambda^*)=\frac{1 + (1-k)\lambda^* + (L + 1)(L + 2 -k)(\lambda^*)^2}{1+\lambda^*},$  $\lambda^*=-1+\sqrt{1+\bar\lambda}.$
\end{prop}

\subsection{How to accelerate the convergence rate?}
We have seen in the previous sections that under suitable conditions one can approximate fixed-points. However,
the major concern associated with the above fixed-point iteration is that the iterates exhibit only linear convergence rate which may be
unacceptably slow.

 Speedup strategic learning is a method  that studies learning mechanisms for speeding up the convergence  based on  few experiences.
The input to a speedup technique typically consists of observations of prior
realized sequence of experience, which may include traces of the measurements of the real problem. The output is knowledge
that the technique can exploit to find solutions more quickly than before and without seriously effecting solution quality.

Our motivation for speedup learning is two-fold:
\begin{itemize}\item
 Traditional results on the fundamental limits of data compression and data transmission through noisy channels apply to the asymptotic regime as codelength, blocklength goes to infinity.  However, these asymptotic results are not usable if the window size and horizon are bounded. Speedup strategic learning is aimed exclusively at finding solutions in
a more practical time frame.
 Therefore, it is interesting to look at the non-asymptotic regime using speedup learning algorithms.
 \item Speedup
 strategic learning aims to create adaptive scheme that can learn patterns from
few number of experience that can be exploited for efficiency gains. Such
adaptive schemes have the potential to significantly outperform traditional
learning schemes by specializing their behavior to the characteristics of the fixed-point problem.
 \end{itemize}

Consider a convergent mean-field learning with exhibit a sequence $\{\bar{m}_t\}_t.$
Suppose now that only few number of  the sequence are available to the users. Each user aims to learn an approximate fixed-point  based only minimal number of information about the estimates $\bar m_0,\ \bar m_1,\ \ldots, \bar m_{T-1}.$
The goal of a generic user is
to accelerate the previous learning algorithm and  transform the slowly converging sequence into a new
one that converge to the exact limit $\bar{m}^*$ as the first one, but faster. If possible, we aim to be as close as possible to $\bar m^*$ based only on the $T$ observations of the sequence $\{\bar m_t\}_t$.

\begin{defi}
Assume that $\bar{m}_t$ converges to $\bar{m}^*$ and let $\eta_t=|\bar{m}_t-\bar{m}^*|.$ If
If two positive constants  $c_1, o>0$  exist, and
$
\limsup_t \ \frac{\eta_{t+1}}{\eta_t^o}=c_1
$

then the sequence $\{\bar{m}_t\}_t$ is said to converge to  $\bar{m}_*$  with order of convergence $o$.  The number  $c_1$  is called the asymptotic error constant.  The cases   $o\in \{1,2\}$ are given special  consideration.

(i)   If  $o=1$ the convergence of  $\{\bar{m}_t\}_t$   is called linear.

(ii)  If  $o=2$ the convergence of  $\{\bar{m}_t\}_t$  is called quadratic.

(iii)  If  $o=3$ the convergence of  $\{\bar{m}_t\}_t$  is called cubic.
\end{defi}

\subsubsection{Partially distributed speedup methods}
We present speedup learning in one-dimensional  space. However,  most of the  speedup schemes below extends to multi-dimensional action spaces.

\subsubsection*{Quadratic order speedup techniques}

The basic Newton method consists to iterate
\begin{eqnarray} \label{eqnewton}
\bar{m}_{t+1}=\bar{m}_t-\frac{g(\bar{m}_t)}{g'(\bar{m}_t)}.
\end{eqnarray}

If $g(\bar{m}^*)=0$ and  $g'(\bar{m}^*)\neq 0$ then the Newton's method generates a quadratic convergence rate locally:
$\eta_{t+1}=|\bar{m}_{t+1}-\bar{m}^*|\leq \frac{h''(\alpha_t)}{2} \eta_{t}^2,$ where $\alpha_t$ is in a neighborhood of $\bar{m}^*$ and $h(y)=y-\frac{g(y)}{g'(y)}.$

For multiple roots, the scheme can be modified to be
$$
\bar{m}_{t+1}=\bar{m}_t-k\frac{g(\bar{m}_t)}{g'(\bar{m}_t)}
$$ or apply the Newton method to the function  $\frac{g}{g'}.$

\subsubsection{Cubic order speedup techniques}
One of the most studied cubic technique is the Halley's method which consists to update as
$$\bar{m}_{t+1}=\bar{m}_t-\frac{2 g(\bar{m}_t)g'(\bar{m}_t)}{2 [g'(\bar{m}_t)]^2-g(\bar{m}_t)g''(\bar{m}_t)}$$

\subsubsection{Arbitrary order speedup techniques}
We start with Householder's speedup method \cite{house}.
If the map $f$ is known by the player then the classical fixed-point iteration is $\bar{m}_{t+1}=g(\bar{m}_t)=f(\bar{m}_t)-\bar{m}_t.$ A $(o+1)-$ order speedup learning is given by
$$
\bar{m}_{t+1}=\bar{m}_t+o\frac{(1/g)^{(o-1)}(\bar{m}_t)}{(1/g)^{(o)}(\bar{m}_t)}
$$
where $o$ is an integer and $(1/g)^{(o)}$ is the derivative of order $o$ of the inverse of the function $g$.

 It is well-known that if $f$ is a $(o+1)$ times continuously differentiable function and $\bar{m}^*$ is a fixed-point of $f$ but not of its derivative, then, in a neighborhood of $\bar{m}^*$, the iterates $\bar{m}_t$ satisfy:
$$| \bar{m}_{t+1}-\bar{m}^*| \leq  c_2 | \bar{m}_t-\bar{m}^* | ^{o+1},$$ for some constant $c_2$ which is obtained by taking the bound of the derivatives of the function $g$ at $m^*.$ The bound is finite because of continuity over compact set.
This means that the iterates converge to the fixed-point if the initial guess is sufficiently close, and that the convergence has rate $(o+1).$ Thus, if $f$ is an infinitely differentiable function, this scheme makes a very fast locally convergent speedup learning algorithm with arbitrary high order. In particular,
for $o=1$ this is the Newton's method, for $o=2$ it is called Halley's method.

For the case where the function $f$ is smooth with a unique fixed-point, we can systematically generate, high order, quickly converging, mean-field learning methods, of any desired degree, for the solution of fixed-point problem.
 Fast converging mean-field learning methods like those should be of great use in large-scale algorithms that require the repetitive solution of a nonlinear equation many times over long time periods  and where an efficient solution algorithm is imperative to avoid overtime computations.
 This reduces the so-called {\it cost of impatience}, i.e., the cost due to the error gap to the solution.
 We now present our main result on the convergence time of the   $(o+1)$-order speedup scheme.
 \begin{thm} \label{labelquick}
 Let $c_2<1$ and $\eta_0=|\bar{m}_0-\bar{m}^*|<1$ and $o>0.$ Then the scheme is convergent and
 the error at iteration $t,$ $\eta_t$ is bounded by
 $$
 \eta_t\leq c_2^{\frac{(o+1)^{t}-1}{o}} \eta_0^{(o+1)^t}
 $$ Thus, the convergence time to an $\eta^*-$range of the pure mean-field equilibrium is
 $$
T_{\eta^*}=1+\lfloor \max(0,T)\rfloor$$ where $$T=\frac{1}{\ln(o+1)}\ \ln\left[ \frac{\ln\left(\frac{1}{\eta^*c_2^{\frac{1}{o}}}\right)}{\ln\left(\frac{1}{\eta_0c_2^{\frac{1}{o}}}\right)} \right]
$$

 \end{thm}

\subsubsection{Derivative-free speedup methods}
The most well known sequence transformations are Aitken's $\Delta^2-$
process (1926,\cite{aitken}), Richardson's extrapolation algorithm (1927, \cite{richardson}), Shanks transformation (1955,\cite{shanks}), Romberg
transformation 1955~\cite{romberg}, Wynn's $\epsilon-$algorithm (1956, \cite{wynn1,wynn2}). These speedup techniques are not based on derivatives.  This is why they are used more frequently in practical computations.

Next we present a superlinear order derivative-free speedup technique, called {\it Secant method}. It is inspired from  Newton's method where the term $ \frac{g(a)-g(a')}{a-a'}$ replaces the derivative.
\subsubsection*{Secant speedup method}
In the Secant speedup method (Algorithm \ref{speedupsecant}), we define the sequence $\bar m_2, \bar m_3, \bar m_4, \ldots$ using two initial guesses, $\bar{m}_0$ and $\bar{m}_1$ and
the formula:
\begin{eqnarray}\label{secanttem}
\bar{m}_{t+1}=\bar{m}_t-\frac{g(\bar{m}_t)(\bar{m}_t-\bar{m}_{t-1})}{g(\bar{m}_t)-g(\bar{m}_{t-1})}
\end{eqnarray}
which can be obtained when replacing $g'(\bar{m}_t)$ by $\frac{g(\bar{m}_t)-g(\bar{m}_{t-1})}{\bar{m}_t-\bar{m}_{t-1}}$ in Equation (\ref{eqnewton}).
Note that the Secant method can be written as $\bar{m}_{t+1}=F(\bar{m}_t,\bar{m}_{t-1})$ which is a two-step memory scheme, since two previous values are required to calculate the
next value in the sequence.
The Secant speedup method converges with order around $1.6,$ i.e., more quickly than a method with linear convergence, but more slower
than a method with quadratic convergence.

 \begin{algorithm}[!htp]
\caption{\textbf{:} Secant speedup method}
\begin{algorithmic}[1]\label{speedupsecant}
\STATE \textbf{Initialization :}\\
Make a starting guess $\bar{m}_{0}, \bar{m}_1$ \\ 

\STATE \textbf{Speedup learning pattern :}\\
for each time slot $t$\\
Observe $\bar{m}_t$ and compute $g(\bar{m}_t)$\\
Compute $\bar{m}_{t+1}=\bar{m}_t-\frac{g(\bar{m}_t)(\bar{m}_t-\bar{m}_{t-1})}{g(\bar{m}_t)-g(\bar{m}_{t-1})}$\\
\end{algorithmic}
\end{algorithm}
\subsubsection*{ Aitken's speedup method}

The new sequence that accelerates the convergence via Aitken extrapolation is
$$
\bar{y}_t=\bar{m}_{t}-\frac{(\bar{m}_{t+1}-\bar{m}_{t})^2}{\bar{m}_{t+2}-2\bar{m}_{t+1}+\bar{m}_{t}},
$$
which is obtained by solving the equation $\frac{\bar{m}_{t+1}-\bar{y}}{\bar{m}_{t}-\bar{y}}=\frac{\bar{m}_{t+2}-\bar{y}}{\bar{m}_{t+1}-\bar{y}}.$

\subsubsection*{Steffensen's speedup method}
Steffensen's speedup method (Algorithm \ref{speedupsteffenson}) is a variant of the Aitken method that uses the Aitken formula to
generate a better sequence directly:

 \begin{algorithm}[htb]
\caption{\textbf{:} Steffensen's speedup method}
\begin{algorithmic}[1]\label{speedupsteffenson}
\STATE \textbf{Initialization :}\\
Make a starting guess ${a}_{0}$ \\ 

\STATE \textbf{Speedup learning pattern :}\\
for each time slot $t$\\
Compute $\bar{m}_1=f(\bar{m}_0),\bar{m}_2=f(\bar{m}_1)$\\
Use Aiken's speedup method to compute $\bar y_0$\\
reStart with $\bar y_0$
\end{algorithmic}
\end{algorithm}

\begin{rem} Sometimes one has attempted to compare Newton, Secant, Aitken and Steffensen speedup methods.
So, which method is faster?
Ignoring constants, it would seem obvious that Newton's method (model-based speedup technique) is
faster than Secant method (non-model speedup technique), since it converges more quickly. However, to compare performance, we must consider both
computational cost and speed of convergence. An algorithm that converges quickly but takes a few seconds per
iteration may take far more time overall than an algorithm that converges more slowly, but takes
only a few milliseconds per iteration. So, the comparison is not fair.
For the purpose of this general analysis, we may assume that the computational cost of an iteration is dominated
by the evaluation of the function. So, the number of function
evaluations per iteration is likely a good measure of cost.
The secant method requires only one function evaluation per iteration (the function $g$), since the value of $g(\bar{m}_{t-1})$
can be stored from the previous iteration.
Newton's method requires one function evaluation and one evaluation of the derivative per
iteration. It is difficult to estimate the cost of evaluating the derivative in general. In some cases,
the derivative may be easy to evaluate, in some cases, it may be much harder to evaluate than the
function (if it is possible at all). If we can run two iterations of the secant speedup method in the time
it will take to run one iteration for computing the derivative in Newton's method. Then, two iterations of the Secant speedup method should be compared to Newton's speedup method. But Secant method with two iterations has a speedup in order of $2o>2,$ hence faster than the Newton's method.

The Aitken speedup method requires three consecutive terms of the sequence $\bar{m}_t$ to reproduce a quadratically convergent speedup technique.
\end{rem}


\subsection{Fully distributed derivative-free mean field learning}
We now present  a fully distributed mean-field learning based on the work of \cite{krstic}. Each generic player adjusts its action based on numerical measurement of own-payoff (with some i.i.d noise).  The first order learning scheme with large number of players is given in Algorithm \ref{algobr3sine}.

\begin{algorithm}[!htp]
\caption{\textbf{:} Mean-field learning with first order sinusoidal perturbation }
\begin{algorithmic}[1]\label{algobr3sine}
\STATE \textbf{Initialization :}\\
for each user $j\in \mathcal{N}$ do\\  
${a}_{j,0},$\\ 

\STATE \textbf{Learning pattern :}\\
for each time slot $t$\\
for each user $j\in \mathcal{N}$ do\\ 
Observe a realized payoff $r_{j,t}$\\
$\hat{a}_{j,t+1}=\hat{a}_{j,t}+\lambda_{j,t}k_j r_{j,t} \epsilon_j\sin(w_j \hat{t}_j+\phi_j)$ \\
$a_{j,t}= \hat{a}_{j,t}+\epsilon_j\sin(w_j \hat{t}_j+\phi_j)$\\
$\hat{t}_j=\sum_{t'=1}^t \lambda_{j,t'}$\\
\end{algorithmic}
where $ \lambda_{j,t}, k_j,\epsilon_j>0,\ \phi_j\in\mathbb{R}.$
\end{algorithm}

The mean  of the mean-field generated by
 Algorithm \ref{algobr3sine} is given by
$$\hat{m}_{j,t+1}=\hat{m}_{j,t}+\liminf_n \ \frac{1}{n}\sum_j \lambda_{j,t}k_j r_{j,t} \epsilon_j\sin(w_j \hat{t}_j+\phi_j),$$
$$m_{j,t}= \hat{m}_{j,t}+\liminf_n \ \frac{1}{n}\sum_j\epsilon_j\sin(w_j \hat{t}_j+\phi_j).$$

Let $\eta_{j,t}:=a_{j,t}-a^*_j.$ From Taylor expansions $\partial_{a_j}r_j(a_t)=\eta_t \partial^2_{a_ja_j}r_j(a^*)+O(|\eta_t|^2)$ and $\partial^2_{a_ja_j}r_j(a_t)=\partial^2_{a_ja_j}r_j(a^*)+O(|\eta_t|).$  In the first order sinusoidal learning, the error rate is proportional to the second derivative (Hessian of the payoff). Since the payoff function is unknown to the players, it is difficult to tune the convergence rate with the appropriate parameters. Therefore, we estimate the Hessian and construct a dynamics that converge asymptotically to the pseudo-inverse of the Hessian. The local behaviour of the second order sinusoidal learning
will be independent of the Hessian which is of big importance in non-model learning where the Hessian is unknown.
The second order sinusoidal perturbation mean-field learning is given in Algorithm \ref{algobr3sine2nd}.

\begin{algorithm}[!htp]
\caption{\textbf{:} Mean-field learning with second order sinusoidal perturbation }
\begin{algorithmic}[1]\label{algobr3sine2nd}
\STATE \textbf{Initialization :}\\
for each user $j\in \mathcal{N}$ do\\  
$\hat{a}_{j,0},$\\ 

\STATE \textbf{Learning pattern :}\\
for each time slot $t$\\
for each user $j\in \mathcal{N}$ do\\ 
Observe a realized payoff $r_{j,t}$\\
$\hat{a}_{j,t+1}=\hat{a}_{j,t}+\lambda_{j,t}k_j \hat{d}_{j,t}^{(2)}r_{j,t} \frac{2}{\epsilon_j}\sin(w_j \hat{t}_j+\phi_j)$ \\
$a_{j,t}= \hat{a}_{j,t}+\epsilon_j\sin(w_j \hat{t}_j+\phi_j)$\\
$\hat{d}_{j,t+1}^{(2)}= (1+\lambda_{j,t}w_c)\hat{d}_{j,t}^{(2)}+\lambda_{j,t} \left[ -w_c \hat{d}_{j,t}^{(2)} s^{(2)}_{j,t} r_{j,t} \hat{d}_{j,t}^{(2)}\right]$\\
$\hat{t}_j=\sum_{t'=1}^t \lambda_{j,t'}$\\
\end{algorithmic}
where $ \lambda_{j,t}, k_j,w_c,\epsilon_j>0,\ \phi_j\in\mathbb{R}.$
\end{algorithm}
 The product of $s^{(2)}_{j,t}r_{j,t}$ needs to generate
an estimate of the Hessian in a time-average  sense and $\hat{d}_{j,t+1}^{(2)}$ should generate an estimate pseudo-inverse of the Hessian. Example of function $s^{(2)}_{j,t}$ is $\frac{1}{\epsilon_j^2}(\sin^2(w_j \hat{t}_j+\phi_j) -1).$

\subsection{Feedback-free mean-field learning}
  Below we  develop a learning algorithm without feedback (no  mean-field feedback, or other actions at the previous step are not observed, \cite{weber}) but with knowledge of the mathematical structure of the payoff function.
Feedback-free mean-field learning is very important and is therefore empirically testable.
In it, players think that others are likely to take the some function of actions as
themselves, resulting in a false consensus or non-false
consensus depending on one's view of the irrationality of the behavior and incompatibility of beliefs and conjectures.

A simple example of feedback-free learning consists to take a speedup version of $\bar m_{t+1}=f(\bar m_t)$ starting from some estimate of the initial point and iterate offline.

\subsection*{Mean-field global optimization}
Consider the average payoff in the form
$\frac{1}{n}\sum_{j\in\mathcal{N}} r(a_j, \bar m_{j,n,t})$ which is in general, not concave in the joint action.
The asymptotic regime is
$\int r(a,\bar{m})\ m(da)$ which can be analyzed using measure theory. In particular, if the integral can be written as a function of only the first moment of the mean-field $m$, i.e.,
$\bar{r}( \bar{m})$ then the above mean-field learning schemes can be used to learn the mean-field social optimum.

\begin{example}
We consider the payoff function in the following form: $r_j(a)=a_j h(\frac{\sum_{i=1}^n a_i}{n} )-p a_j.$ Then total payoff of all the players is
$\sum_{j=1}^n r_j(a)=D h(D/n) -p D$ where $D$ is the total sum of actions. By dividing by $n,$ one gets $$\frac{1}{n} \sum_{j=1}^n r_j(a)=\frac{D}{n} h(D/n) -p \frac{D}{n}=\bar{m}_n h(\bar{m}_n)-p \bar{m}_n$$
The dimensionality of the global optimization problem can be significantly reduced to be
one-dimensional, i.e., It suffices to optimize the function $z h(z)-p z.$ The local extrema of this scalar function can be found using the above techniques.

\end{example}

This means that a mean-field optimization can be conducted easily for any limiting function in the form of $a h(\bar{m})+\beta_1 a+\beta_2 g(\bar{m})+\beta_3$ where $\beta_i$'s are real numbers and $h,g$ are limiting functions of the mean of the mean-field.

    In next section, we apply mean-field learning and speedup techniques in beauty contest game or guessing game.

\section{Beauty contest game} \label{exsec} We revisit the beauty contest game in the context of mean-field.
The name of this game and its basic idea go back to John Maynard Keynes (1936) who compared a clever investor to a participant in a newspaper beauty-contest where the aim was to guess the average preferred face among 100 photographs.
The initial beauty contest game was analyzed for integers,
although in 1993 the German economist Rosemarie Nagel based her experiments on a nice variant of the game, played by Keynes' newspaper readers:
Each player chooses a real number between 0 and 100 inclusively. The number need not be an integer. A player wins if its number is closest to 2/3 of the average of the numbers given by all participants.

There are two ways to see that a unique equilibrium solution exists. First, one can easily see that no one should submit a number higher than 66, because whatever the others do, a guess higher than 66 cannot be better than 66. However, if no one guesses more than 66, then all numbers between 44 (that is, two third of 66) and 66 are inferior to 44. Hence no one would guess more than 44 and so on.
until 0 is the only remaining reasonable choice. (Note that a second solution besides 0 is 1 if only integers are allowed).

A second method would be to just try out: Presume you guess a certain number $a$, and anyone else guesses the same number $a$, would you still wish to stick to your initial guess $a$? If so, you have found  a symmetric Nash equilibrium. Now the only $a$ to which you would want to stick is $0.$

More generally, in a beauty contest game, $n\geq 3$ players simultaneously choose numbers $a_j$ in
some interval, say, $[0,M],\ M>0.$ The average of their numbers $\bar{m}_n=\frac{1}{n}\sum_ja_j$ is computed,
which establishes a target number $p\bar{m}_n,$ where $0<p<1$ is a parameter. The player whose
number is closest to the target $p\bar{m}_n$ wins a  prize $\alpha_n.$ The generic payoff is

$$
r_j(a)=\alpha_n \frac{\ind_{\{ a_j\in \arg\min_{a'_j} |a_j'-p\bar{m}_n|\}}}{\sum_{i} \ind_{\{a_i\in \arg\min_{a'_i} |a_i'-p\bar{m}_n| \}}}
$$

This model of beauty contest games were  studied experimentally by Nagel (1995, \cite{nagel95}).

These games are useful for estimating the number of steps of iterated dominance
players use in reasoning through games. To illustrate, suppose $p<1$.
Since the target can never be above $pM$, any number choice above $pM$ is
stochastically dominated by simply picking $pM$. Similarly, players who obey
dominance, and believe others do too, will pick numbers below $p^2 M$ so choices in
the interval $(p^2M,M]$ violate the conjunction of dominance and one step of
iterated dominance. We iterate this progressively and get that $p^{t}M\longrightarrow 0$, then the unique Nash equilibrium is 0.
Now, if $p>1$ then the equilibrium is $M.$ If $p=1$ every feasible symmetric action profile is an equilibrium.

We now consider a small modification of the guessing game in order to get interior equilibria.
The target is changed to be
$\mu+p\bar{m}_n, \ \mu\geq 0$
The generic payoff is
$$
r_j(a)=R -\kappa \left\Vert a_j-\left(\mu+\frac{p}{n}\sum_{j'=1}^n a_{j'} \right) \right\Vert,\
$$
where $(R\geq 0, \kappa>0, \ \mu\geq 0).$

In the asymptotic regime, the interior mean-field equilibrium (if any) should satisfy $\bar{m}=\mu+p\bar{m}, p<1, $ i.e., $\bar{m}^*=\frac{\mu}{1-p}$ in the interval $[0, M].$

Assuming that the functions are known, we can use a feedback-free mean-field learning. Therefore, one possible explanation of no-feedback learning in the beauty contest game (or guessing game) is that players simply take an
action, treat their action as representative of the choices of other players, and then best
respond to this belief mean of the mean-field. This kind of reasoning would predict convergence towards equilibrium
in the  guessing game.

The best response dynamics is given by $\min (M, \mu+p \bar{m}_{t-1})$ where the starting point is $\bar{m}_0\geq 0.$
Each trader starts with an estimate $\hat{m}_0.$ It is clear that for $\mu=0,p<1$ if each trader estimates the initial point and iterate the mean-field learning process offline, the process  converges to the mean-field equilibrium. For $\mu>0,$ the interior response writes $\bar m_{t+1}=\mu+p\bar m_{t}.$

More generally, one can consider a mean-field payoff in the following form
$$
r_j(a_j, \bar m)=R -\kappa \|a_j-\chi(\bar m) \|,\
$$ where $\chi$ is map which has a fixed point in $[0, M].$
The best response to the mean $\bar m$ is $\chi(\bar m)$ and the mean-field pure equilibrium satisfies
$a^*=\bar m^*=\chi(\bar m^*).$ Hence, learning the mean-field equilibrium reduces to learning a fixed-point of the maping $\chi.$ For  $\chi(\bar m)=\sqrt{2\bar m+3}$ the iterative process becomes $\bar{m}_{t+1}=\sqrt{3+2\bar{m}_{t}}.$
Note that $3$ is a fixed-point. Let fix the starting point at $\bar{m}_{0}=4$ and $M=100.$

 We start with $g(\bar{m})=\bar{m}^2-2\bar{m}-3$ and use the secant speedup method. The result of acceleration technique (\ref{secanttem}) is presented in Table \ref{table2tret1}. We observe that at the fourth iteration, the secant speedup method has already $10^{-4}$ of precision while the original sequence has a $100$ times smaller precision.  Next we start with the initial point $5$ for secant method and initial point of $4$ for the fixed point method. We observe that the secant method has better precision than the fixed-point method after only $3$ iterations. This means that the secant speedup method is robust to  initial estimation errors.
\begin{table}[htb]
  \centering
  $
  \begin{array}{ccc}
    Original\ sequence & Secant \ speedup\  method &     \\
    \bar{m}_0 = 4 &  4 & 5\\
    \bar{m}_1 = 3.316624790 &  3.3166 & 3.6056  \\
    \bar{m}_2 =3.103747667 &  3.0595 &  3.1833\\
    \bar{m}_3 =3.034385495 &   3.0043 & 3.0232 \\
    \bar{m}_4 = 3.011440019&  3.0001 & 3.0010
  \end{array}
  $
  \caption{Fixed-point and Secant speedup method}\label{table2tret1}
\end{table}

We summarize the acceleration technique in Table \ref{table2tret} based only on few steps of  the original sequence.

\begin{table}[htb]
  \centering
  $
  \begin{array}{ccc}
    Original\ sequence & Aitken     &  Steffensen \\
    \bar{m}_0 = 4 & 3.007431293 &  3.000000510\\
    \bar{m}_1 = 3.316624790 & 3.000862083  &  3.000000000000002\\
    \bar{m}_2 =3.103747667 & 3.000097228  & \\
    \bar{m}_3 =3.034385495 &  & \\
    \bar{m}_4 = 3.011440019&  &
  \end{array}
  $
  \caption{Acceleration of mean-field learning}\label{table2tret}
\end{table}

Assume that only $5$ measurements of the mean sequence is given to the player:  $\bar{m}_0=4,$ $\bar{m}_{1}=3.316624790,\bar{m}_{2}=3.103747667,\bar{m}_3:=3.034385495 , \bar{m}_4=3.011440019.$
We apply the acceleration technique from these  measurements and one gets
$
\bar{y}_0=\bar{m}_{0}-\frac{(\bar{m}_{1}-\bar{m}_{0})^2}{\bar{m}_{2}-2\bar{m}_{1}+\bar{m}_{0}}= 3.007431293
$

$
\bar{y}_1=3.000862083
$
and $\bar{y}_2=3.000097228$

\begin{figure}
\centering
  \includegraphics[width=0.9\columnwidth]{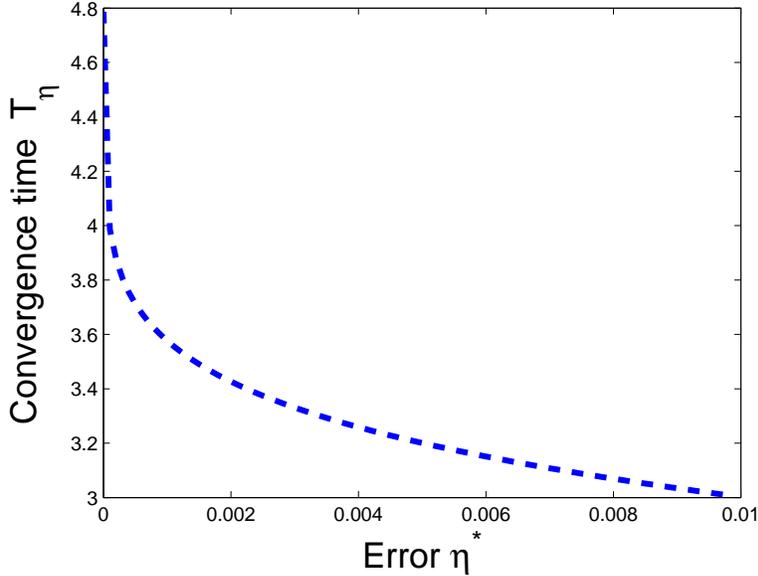}\\
  \caption{Illustration of the convergence time  bound from Theorem \ref{labelquick}. 5 iterations provide already remarkable result. }\label{figtime}
\end{figure}

Clearly, the Aitken sequence $\{\bar{y}_t\}_t$ guarantees that will
converge faster to $3$ and the error  will be
smaller than that of the original sequence $\{\bar{m}_t\}_t.$
If we reiterate the fixed-point of the sequence but starting from the sequence $\bar{y}_0.$ Then,
$
\bar{z}_0=\bar{y}_{0}-\frac{(\bar{y}_{1}-\bar{y}_{0})^2}{\bar{y}_{2}-2\bar{y}_{1}+\bar{y}_{0}}=
3.000000510.
$
Repeating the acceleration procedure and  taking the fixed-point iteration, one gets
$\bar{z}_1=3.000000000000002.$ The sequence $\{\bar{z}_t\}$ seems converges to $3$ with at least quadratic convergence rate which is great acceleration from the linear convergence  rate of the original sequence.
With only two iterations of Steffensen's speedup method one has reached an error of $10^{-6}$ which is satisfactory.
In Figure \ref{figtime} we illustrate the bound of  Theorem \ref{labelquick} which states the convergence time with  error $\eta^*=10^{-4}.$ We observe that the result of Figure \ref{figtime} are similar to the one obtained in Table \ref{table2tret}.

\section{Speedup strategic learning  for satisfactory solution} \label{secsat}
One of the fundamental challenges in distributed interactive systems is to design efficient and fair solutions.  In such systems, a satisfactory solution is an innovative approach that aims to provide all  players with a satisfactory payoff anytime anywhere.
 Our motivations for satisfactory solution seeking are the following: In dynamic interactive system, most users constantly make decisions which are simply ``good enough'' rather than best response or optimal. Simon (1956, \cite{simon})  has adopted the word ``satisficing'' for this  type of decision. Most of literature of strategic learning and decision making problems, however, seek only the optimal solution or Nash equilibria based on rigid criteria and reject others.  As   mentioned by Simon himself in his paper in  page 129,   {\it ``Evidently, organisms adapt well enough to 'satisfice'; they do not, in general, 'optimize'.} Therefore satisfactory  solution offers an alternative approach and is closely model the way humans and cognitive users make decisions~\cite{wynnt,wynn2t,wynn3t}.

Here, a satisfactory strategy is a decision-making strategy that attempts to meet an acceptability threshold. This is contrasted with optimal decision-making or best response strategy, an approach that specifically attempts to find the best option available given the the choice of the other users. Following that idea we define a satisfaction solution as a situation where every user is satisfied, i.e., her payoff is above her satisfactory level.

In this section we focus on fully distribution strategic learning for satisfactory solution in games with continuous action space. For discrete (and finite) action space we refer to \cite{sp}. We show that the methodology in \cite{sp} can be extended to continuous action space as well as to first moment mean-field games. We illustrate it with a basic example. See  also \cite{accpedro}.

\begin{defi}
The action profile $(a_j)_{j\in\mathcal{N}}$ is a pure satisfactory solution of the game $\mathcal{G}$ if all the players are satisfied:
$$
\tilde{r}_j(a)\geq \gamma^*_j,\ \forall j\in\mathcal{N}
$$
\end{defi}

Before going for pure satisfactory solution seeking we first need  to ask if the problem is well-posed, i.e; the existence of a pure satisfactory solution.

We assume  feasibility, that is, $\gamma^*_j,\ j\in\mathcal{N},$ are chosen such that the set $\{ a=(a_1,\ldots,a_n)\in \mathcal{A}^n\ | \ \tilde{r}_j(a)\geq \gamma_j^*,\  \forall j \}$ is nonempty.
This means that there exists  a vector $(\epsilon_1,\ldots,\epsilon_n),$ $\epsilon_j\geq 0$ such that there is an action profile $a$ that satisfies
$\forall j, \  \tilde{r}_j(a)=\gamma^*_j+\epsilon_j.$ Thus, a necessary and sufficient condition for existence of a satisfactory solution is that the vector $\gamma^*+\epsilon$ belongs to the set $\hat{r}(\mathcal{A}^n)$ i.e., the range of the function $\hat{r}(a):=(\tilde{r}_1(a),\ldots, \tilde{r}_n(a)).$

Consider a basic wireless network with $n$ users. The action space is $\mathcal{A}_j=[0, a_{j,\max}],\ a_{j,\max}>0.$ There is a state space $\mathbb{R}^{n\times, n}_{+},$ $w=(w_1,w_2,\ldots,w_n)$ where $w_j=(w_{jj'})_{j'},$ $w_{jj'}=|h_{jj'}|^2 \geq 0, \ h_{jj'}\in\mathbb{C}.$
The payoff of user $j$ in state $w$ is the signal-to-interference-plus-noise ratio: $r_j(w,a)=SINR_j(w,a)=\gamma_j(w,a)=\frac{a_jw_{jj}}{N_0+\sum_{j'\neq j}a_{j'}w_{jj'}\epsilon_{jj'}}$ where $N_0>0$ is the background noise and $\epsilon_{jj'}>0.$ To goal of each player is not necessarily to maximize the payoff, it is to get a certain target $\gamma^*_j.$
\begin{defi}We say that user $j$ is {\it satisfied}  if  $\gamma_j(w,a)\geq \gamma^*_j.$
\end{defi}
A satisfactory solution at state $w$ is a situation $a$ where all the users are satisfied. Such a situation may not exist in general. We examine the case where there is at least one solution.
 Of course if the full state $w$ and all the parameters are known, one can perform a centralized solution. However, in the distributed setting, a user may not have access to the information of the other users channel gains and their locations. Thus, it is important   to guarantee a certain quality-of-service (QoS) with minimal information for all the users.
 Our goal here is to develop  very fast and convergent fully distributed learning algorithms for  satisfactory solutions. The only information assumption required to each user is the numerical realized value its own-payoff $r_{j,t}$ and its own-target $\gamma^*_j.$ The basic Banach-Picard fixed-point iteration is given by
$$
a_{j,t+1}=\mbox{proj}_{\mathcal{A}_j}\left[a_{j,t}\frac{\gamma^*_j}{r_{j,t}}\right],
$$
where $\mbox{proj}_{\mathcal{A}_j}$ denotes the projection operator over the convex and compact set $\mathcal{A}_j,$ i.e., $\mbox{proj}_{\mathcal{A}_j}(x)=\min(a_{j,\max}, \max(0,x)).$

Note that $\max(0,x)=[x]_+=\frac{x+|x|}{2},$ and $\min(a_{j,\max}, \alpha)=\frac{a_{j,\max}+\alpha-| a_{j,\max}-\alpha|}{2}.$

The proposed algorithm is fully distributed in the sense that a user do  not need to observe the actions of the others in order to update its strategy iteratively.
 \begin{algorithm}[!htp]
\caption{\textbf{:} Fully distributed Banach-Picard learning for satisfactory solution}
\begin{algorithmic}[1]\label{deter}
\STATE \textbf{Initialization :}\\
Make a starting guess ${a}_{j,0}$ \\ 

\STATE \textbf{Banach-Picard learning pattern :}\\
For each time slot $t$ up to $T$\\
For each user $j\in \mathcal{N}$ do:\\   
Observe a numerical value $r_{j,t}$ \\
Compute $a_{j,t+1}=\mbox{proj}_{\mathcal{A}_j}\left[a_{j,t}\frac{\gamma^*_j}{r_{j,t}}\right]$\\
\end{algorithmic}
\end{algorithm}

Next we discuss the convergence of the Banach-Picard learning algorithm for a fixed state $w.$

{\bf Assumption A0: } $\rho(M^w)<1$ where $M^w_{j'j}=\frac{w_{jj'}\gamma_j^*}{w_{jj}}$ and $M^w_{jj}=0.$

It is clear that under assumption A0, the system $(I-M^w)a=b$ where $b_j=\frac{\gamma_j^*N_0}{w_{jj}}$ has a solution.
We say that the problem is feasible if  $(I-M^w)a=b$ has a solution and the solution $a^*$ satisfies $0<a_j^*\leq a_{j,\max}.$

\begin{prop} \label{feasiblethm}Consider the Banach-Picard learning algorithm for a fixed state $w$ for which the problem is feasible.
\begin{itemize}
\item Suppose that  the sequences of action profiles $\{a_t\}_t$ generated by the Banach-Picard algorithm converges to some point $a^*$ that belongs to be relative interior of $\prod_j \mathcal{A}_j.$ Then, $a^*$ is a satisfactory solution.
    \item Under the sufficient condition for existence A0 and feasibility condition, the Banach-Picard iteration converges to a satisfactory solution.
\item
The convergence rate of the Banach-Picard algorithm for satisfactory solution seeking is geometrical decay and hence the convergence time within $\eta$ error tolerance is
$$
T_{\eta}= 1+\lfloor \max(0,T)\rfloor$$ where $$T=\frac{\ln [ \frac{d(a_0, a^*)}{\eta }]}{\ln \frac{1}{\rho(M^w)}}
$$
\end{itemize}
\end{prop}

\begin{rem}[Advantages]
This proposition is very important since a satisfactory solution can be seen as a global optimum of the game with payoff  function $\ind_{\{ r_j\geq \gamma_j^*\}},$ where $\ind_{\{.\}}$ denotes the indicator function. In particular, the above algorithm is a fully distributed learning scheme that converges (under the existence and feasibility condition) to a global optimum (and hence Pareto optimal) which is remarkable.
\end{rem}

\begin{rem}[Limitations]
 The fully distributed Banach-Picard algorithm proposed above is convergent under some  range of parameters, and the algorithm is with minimal information (it is fully distributed). However, the convergence time is still unsatisfactory. We aim to investigate whether it is possible to get a faster convergence rate. To do so, we use speedup learning techniques.
 \end{rem}

 One of the first speedup techniques for satisfactory solution is
 the reverse Ishikawa's  learning consists to choose a learning rate (bigger that one) that converges to one.
\begin{eqnarray}
a_{t+1}=\mbox{proj}_{\mathcal{A}_j}\left[\lambda_t a_{j,t}\frac{\gamma^*_j}{r_{j,t}}+(1-\lambda_t)a_{j,t}\right] \\
1<\lambda_t< 2,\ \lim_t \lambda_t=1,\
a_0\in \mathcal{A}.
\end{eqnarray}
\begin{thm} \label{feasiblethm2}
Under the same assumption as in Proposition \ref{feasiblethm} and appropriate choice of $\lambda_t,$ the reverse Ishikawa learning converges faster than the Banach-Picard learning.
\end{thm}

Note that the projection is now required even if $\mathcal{A}_j$ is convex because for $\lambda_t>1,$ one gets $1-\lambda_t<0$ is a not convex combination. In general, it is difficult to compute in advance the value of  $\lambda_t$ that will maximize the rate of convergence.

In order to get a higher order convergence rate, one can use a Steffensen speedup learning of the reverse Ishikawa.

\begin{rem}
In the above speedup learning for satisfaction, we have limited ourselves to the case where the state is quasi-static. However, in wireless networks, it could be stochastic, leading to a stochastic learning algorithm. Then, the goal is to find  satisfactory solution in expectation. For that case, the assumption on $M^w$ is too restrictive. The spectral radius of the  matrix $M^w$ may not be less than one for some realized state $w.$ Then, a feasible solution may not exist but the algorithm converges to $a_{j,\max}.$
\end{rem}

In the context of large-scale games, one needs to scale the SINR. By choosing the parameter $\epsilon,$ and $a_{j,\max}=\bar{a}_{\max}>0,\gamma_j=\gamma^*$ independent of $j$ one can express the payoff of a generic user as function of $a_j,$  $\bar{m},$ and a load factor $\alpha.$ Therefore, the mean-field learning version becomes
\begin{eqnarray}
\bar m_{t+1}&=&\mbox{proj}_{\bar{\mathcal{A}}}\left[\bar m_{t}\frac{\gamma^*}{r_{t}}\right]\\
&=&\mbox{proj}_{\bar{\mathcal{A}}}\left[\gamma^*(\bar{N}_0+\alpha \bar m_t)\right] \\
&& \bar m_0\in \bar{\mathcal{A}}=[0,\bar{a}_{\max}].
\end{eqnarray}
Thus, $\rho(M^w)=\gamma^*\alpha<1$ and $\bar m^*=\frac{\bar{N}_0\gamma^*}{1-\gamma^*\alpha} <\bar{a}_{\max}$ is an interior satisfactory solution and the payoff of a each generic player is $ \frac{\bar m^*}{\bar{N}_0+\alpha \bar{m}^*}=
\frac{\bar m^*}{\bar{N}_0 (1+\frac{\alpha\gamma^*}{1-\gamma^*\alpha})}=\frac{\bar m^*(1-\gamma^*\alpha)}{\bar{N}_0}=\gamma^*
$

 The reverse Ishikawa is
\begin{eqnarray}
\bar m_{t+1}& =&\mbox{proj}_{\bar{\mathcal{A}}}\left[\lambda_t \bar m_{t}\frac{\gamma^*}{r_{t}}+(1-\lambda_t)\bar m_{t}\right]\\ &=&
\mbox{proj}_{\bar{\mathcal{A}}}\left[\lambda_t\gamma^*(\bar{N}_0+\alpha \bar m_t)+(1-\lambda_t)\bar m_{t}\right]\\
&=&  \mbox{proj}_{\bar{\mathcal{A}}}\left[\lambda_t\gamma^*\bar{N}_0 + (\lambda_t\gamma^*\alpha+(1-\lambda_t)) \bar m_t\right]
 \\
&& 1<\lambda_t< 2,\
\bar m_0\in \bar{\mathcal{A}}.
\end{eqnarray}
For $\lambda_t=\lambda\in (1,2),$ one can observe compare the spectral radius:
$0<\lambda\gamma^*\alpha+(1-\lambda) < \gamma^*\alpha <1$ Thus, the reverse Ishikawa learning has a superlinear convergence rate faster than the Banach-Picard fixed point iteration.

We choose the target SINR to be $\gamma^*=20,$ the scaled background noise $\bar{N}_0=0.3,$ and the load is $\alpha=1/30$  and $\bar{a}_{\max}=20.$ Then, the problem is feasible and the satisfactory solution is $18.$
In table \ref{table3satisfaction} we illustrate the convergence to satisfactory solution: Banach-Picard and its speedup versions with reverse Ishikawa and Steffensen. We initialize the mean of the mean field $\bar{m}_0=2$ and observe that $50$ iterations of the Banach-Picard learning corresponds approximately to 25 iterations of  the reverse Ishikawa learning and only $5$ iterations of Steffensen speedup algorithm.
\begin{figure}[h]
  \centering
  \includegraphics[width=0.9\columnwidth]{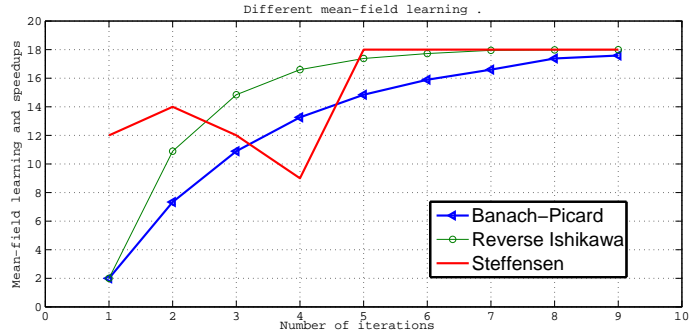}
  \caption{Mean-field learning for satisfactory solution: Banach-Picard, reverse Ishikawa and Steffensen.}\label{plotemfig1}
\end{figure}
Figure
 \ref{plotemfig1} summarizes the three mean-field learning trajectories.

 \begin{table}[htb]
  \centering
  {\tiny
  $
  \begin{array}{|c|c|c|c|} \hline
     t &        Banach-Picard   \      m_t & \lambda=5/3 & Steffensen\\
   1  &  2.000000000000000 &   2      & 12\\
   2  & 7.333333333333333 &    10.888888888888888      & 14\\
   3 &  10.888888888888888 &    14.839506172839506     & 12.000000000000002\\
   4 & 13.259259259259260 &      16.595336076817560    & 9.000000000000002\\
   5 &  14.839506172839506 &   17.375704923030028                     & 17.999999999999986\\
   6& 15.893004115226336 &   17.722535521346678               & 17.999999999999979\\
   7&  16.595336076817556 &17.876682453931856 &\\
   8& 17.063557384545035 & 17.945192201747496 &\\
   9&  17.375704923030021 & 17.975640978554445 &\\
  10&  17.583803282020014 & 17.989173768246424 &\\
  11&  17.722535521346675 & 17.995188341442855 &\\
  12&  17.815023680897784 & 17.997861485085711 &\\
  13& 17.876682453931856 & 17.999049548926983 &\\
  14&  17.917788302621240 & 17.999577577300883 &\\
  15&  17.945192201747492 & 17.999812256578174 &\\
  16&  17.963461467831664 & 17.999916558479192 &\\
  17&  17.975640978554441 & 17.999962914879639 &\\
  18&  17.983760652369625 & 17.999983517724282 &\\
  19&  17.989173768246417  & 17.999992674544124 &\\
  20&  17.992782512164279 & 17.999996744241830 & \\
  21&  17.995188341442852 & 17.999998552996367  &\\
  22&  17.996792227628568 & 17.999999356887276 &   \\
  23&  17.997861485085714 & 17.999999714172120 &   \\
  24&  17.998574323390475 & 17.999999872965383 &   \\
  25&  17.999049548926983 & 17.999999943540168 &    \\
  26&  17.999366365951325 & 17.999999974906743 &   \\
  27&  17.999577577300883 &   & \\
  28&  17.999718384867258 &   &  \\
  29&  17.999812256578171 &   &   \\
  30&  17.999874837718778 &  &     \\
  31&  17.999916558479185 & & \\
  32&  17.999944372319455 & &  \\
  33&  17.999962914879639 & & \\
  34&  17.999975276586426 & &\\
  35&  17.999983517724285 & &\\
  36&  17.999989011816190 &  &\\
  37&  17.999992674544124 & & \\
  38&  17.999995116362747&   &\\
  39&  17.999996744241834 &  &\\
  40&  17.999997829494557 &  &\\
  41&  17.999998552996370 &  &\\
  42& 17.999999035330912  &  &\\
  43&  17.999999356887276 &  &\\
  44&  17.999999571258183 &  &\\
  45&  17.999999714172120 &  &  \\
  46&  17.999999809448077 &  & \\
  47&  17.999999872965383 &  & \\
  48&  17.999999915310255 &  & \\
  49&  17.999999943540171 &  & \\
  50&  17.999999962360114 &  &  \\ \hline
\end{array}
  $}
  \caption{Acceleration of mean-field learning for Satisfactory solution.}\label{table3satisfaction}
\end{table}

The satisfactory solution estimated   by Banach-Picard mean-field is $\bar{m}^* =17.999999962360114$ and the
the error estimate for $\bar{m}^*$ in the Banach-Picard learning is  $ 1.2547\times 10^{-8}$ after $50$ iterations.
The first speedup technique based on reversed Ishikawa with $\lambda=5/3>1$ gets $17.999999974906743$ after $26$ iterations which clearly a superlinear convergence rate.  The error estimate of reverse Ishikawa speedup is  $ 1.3941 \times 10^{-8}$ after $26$ iterations starting from $\bar{m}_0=2$ which is far away from the satisfactory solution. We can get a higher order convergence rate.
The speedup technique \`a la Steffensen provides an error of $7\times 10^{-15}$ after only $6$ iterations. The numerical gap $d(\bar{m}_t, f(\bar{m}_t))$ is in order of $
    7.105427357601002 \times 10^{-15}$ which is an acceptable error tolerance for 6 iterations.

\section{Concluding remarks} \label{cont}


In this paper we have studied mean-field learning in large-scale systems. Our result shows that in large-scale aggregative games with additive aggregation term, the mean-field learning simplifies not only  the complexity of the problem (instead of systems of iterative equations, we can just reduce to one-equation per class or type) but  also the information requirement. We have examined both convergence time and speed of convergence and proposed acceleration techniques for partially distributed mean-field learning with convergence time  in order of $O(\log(\log \frac{1}{\eta})).$

In the fully distributed mean-field learning,
we have assumed that each player is able to observe/measure a noisy (numerical) value of her payoff in a linear way. Now, what happens if this assumption does not hold but a correlated non-linear signal is observed as output?
After an experimentation at time $t,$ player $j$ observes the realized process $O_{j,t}=\tilde{f}(r_{j,t},\xi_{j,t})$ where $\tilde{f}$ is the observation function and $\xi_{j,t}$ is the measurement noise.
If the observation function $\tilde{f}$ is binary then one gets a $0-1$ output or a noisy ACK/NACK feedback.
If $\tilde{f}$ is known and invertible with the respect to the first component, one can use a {\it non-linear mean field estimator} to track the ``true'' payoff function simultaneously with the strategy.
As a third alternative, we have seen that no-feedback mean-field learning is possible. In it, each player estimates the initial mean and iterates offline without any observation/signal from the system.
Several questions remain open:

(i)  In the mean-field learning without feedback,
How to estimate  the starting point by each player and what is the impact on the inconsistency of the process with the respect to the mean-field?

 (ii) What is the outcome of the mean field game if  some fraction of players are with partially distributed learning, some fraction with fully distributed learning schemes and some others without any feedback learning?

(iii) How to extend the mean-field learning framework to  payoff functions that depend not only on the mean also but on higher moments or the entire mean-field distribution?

(iv) Our analysis of speedup learning algorithms are limited to deterministic function. It is interesting to investigate the stochastic version of (\ref{deter}).

We do not have  answers to these questions and postpone them for future investigation.

\appendix

{\bf Proof
of Theorem \ref{labelquick}:}
Let $t\geq 2.$ We reiterate the recursive equation
\begin{eqnarray}
\eta_t &\leq & c_2 \eta_{t-1}^{o+1}\\
 & \leq &   c_2 \left( c_2\eta_{t-2}^{o+1}\right)^{o+1}=c_2^{1+o+1}\eta_{t-2}^{(o+1)^2}\\
 & \leq & c_2^{1+(o+1)}\left( c_2\eta_{t-3}^{o+1}\right)^{(o+1)^2}\\ &=& c_2^{1+(o+1)+(o+1)^2}\eta_{t-3}^{(o+1)^3}\\
&\leq & c_2^{1+(o+1)+(o+1)^2+\ldots+(o+1)^{t-1}}\eta_{0}^{(o+1)^t}
\end{eqnarray}
We remember that for $q\neq 1,\ $ $1+q+\ldots+q^{t-1}=\frac{q^{t}-1}{q-1}.$  Thus,
$$
\eta_t\leq  c_2^{\frac{(o+1)^t-1}{o}}\eta_{0}^{(o+1)^t}.
$$
This means that the convergence time to be within an $\eta^*-$neighorbhood of the mean-field equilibrium is at most for $t$ satisfying
$c_2^{\frac{(o+1)^t-1}{o}}\eta_{0}^{(o+1)^t}\leq \eta^*.$
Thus,
$$(\eta_0c_2^{\frac{1}{o}})^{(o+1)^t}\leq  \eta^*c_2^{\frac{1}{o}}.$$
By taking the logarithm twice, one gets
$$
T_{\eta^*}=\frac{1}{\ln(o+1)}\ \ln\left[ \frac{\ln\left(\frac{1}{\eta^*c_2^{\frac{1}{o}}}\right)}{\ln\left(\frac{1}{\eta_0c_2^{\frac{1}{o}}}\right)} \right]
$$
which is the announced result.

{\bf Proof of Proposition \ref{proth1}:}

Using the strict contraction map $f,$ we estimate the error at time $t$ and get

$\frac{\alpha_1^t}{1-\alpha_1}d(\bar{m}_0,\bar{m}^*)\leq \eta$

i.e. $$\alpha_1^t\leq \frac{\eta (1-\alpha_1)}{d(\bar{m}_0,\bar{m}^*)}.$$ Taking the logarithm yields,
 $$T_{\eta}=\frac{\ln [ \frac{d(\bar{m}_0, fix(f))}{\eta (1-\alpha_1)}]}{\ln \frac{1}{\alpha_1}}.$$ This completes the proof.

{\bf Proof of Proposition \ref{th2time}:}

To prove the convergence time of learning with non-expansive map, we use the Theorem \ref{thmbaillon}. The error bound to an approximated fixed-point is
$$2\frac{d(\bar{m}_{0}, fix(f))}{\sqrt{\pi \sum_{t'=1}^t\lambda_{t'}(1-\lambda_{t'})}}.$$  Remark that $\lambda(1-\lambda)\leq \frac{1}{4}$ for $\lambda\in [0,1].$ Then, the convergence time is at most for a time $t$ that satisfies
$$4\frac{d(\bar{m}_{0}, fix(f))}{\sqrt{\pi t}}\leq \eta$$
$$
\frac{16 d(\bar{m}_0, fix(f))^2}{\eta^2\pi}\leq t.
$$
Hence,
$$T_{\eta}=\frac{16 d(\bar{m}_0, fix(f))^2}{\eta^2\pi}$$

{\bf Proof of Proposition \ref{proppseudo}:} The proof follows immediately from the geometric decay inequality
in Theorem \ref{pththm3} following similar lines as in  Proposition \ref{proth1}.

{\bf Proof of Proposition \ref{abconv}:}
Let $z_t=a_{\int_0^t \lambda_s\ ds}.$  The function $z$ is differentiable and the time derivative is
$$\dot{z}_t=f({a}_{\int_0^t \lambda_s\ ds}). \frac{d}{dt}\left[ \int_0^t \lambda_s\ ds\right]=\lambda_t f({a}_{\int_0^t \lambda_s\ ds})=\lambda_t f(z_t).$$ Moreover, $z_0=a_0.$ By Cauchy's theorem, $z_t=b_t.$
Suppose now that $a_t$ reaches a target set $O$ with at most $T_a$ time units. Then, the trajectory of $b$ reaches the same set $O$ for at most $T_b=g^{-1}(T_a)$ where $g:\ t\longrightarrow \int_0^t \lambda_s\ ds.$ Since if $\lambda\geq 0$ and $\lambda$ non-integrable then $g(t)=T_a$ has a solution.

If $\lambda_s=\lambda,\ $ then $T_b=\frac{T_a}{\lambda}.$

If $\lambda_s=e^{s},\ $ then $T_b=\ln(T_a+1),$ i.e., $\frac{T_b}{T_a}$ goes to zero and $b$ is faster than $a.$

{\bf Proof of Proposition \ref{feasiblethm}: }

If the algorithm converges to a some point $a^*$ then combining the continuity of the projection map and the continuity of  the payoff function,  one gets the righthandside of
$$a_{j}^*=\mbox{proj}_{\mathcal{A}_j}\left[a_{j}^*\frac{\gamma^*_j}{r_{j}^*}\right]$$ which is continuous in $a_t.$ Taking the limit as $t$ goes to infinity, one gets
 $a_j^*=a_{j,\max}$ or $a^*=(I-M^w)^{-1}b$ and the payoff of player $j$ is $r_j^*=\gamma^*_j$ which means that every player $j$  is satisfied in interior steady state.

Now assume  feasibility and assumption A0. Then, From Perron-Frobenius theorem, we known that
$(I-M^w)^{-1}$ exists and
$(I-M^w)^{-1}b$ is positive componentwise. Under feasibility condition, the algorithm generates an error
as
$$
d(a_t,a^*)\leq \rho(M^w)^t d(a_0,a^*),
$$ which provides the convergence of the algorithm to $a^*.$
We use similar analysis as in Proposition \ref{proppseudo} to deduce the convergence time
$$
T_{\eta}= \frac{\ln [ \frac{d(a_0, a^*)}{\eta }]}{\ln \frac{1}{\rho(M^w)}}
$$

{\bf Proof of Theorem \ref{feasiblethm2}: }

Following the proof of Proposition~\ref{feasiblethm2}, one gets that there is some time $T\geq 1$ such that for all  $t\geq T$ the spectral radius of the time-varying matrix $\lambda_t M^w+(1-\lambda_t)I$ for $\lambda_t\in (1,2)$ is less that $\rho(M^w)$ i.e., the reversed Ishikawa has a superlinear convergence rate. If $\rho(M^w)<1$ then both algorithms converge to the same point and the reverse Ishikawa learning algorithm converges faster than the Banach-Picard fixed-point. The reverse Ishikawa is a speedup version of the Banach-Picard algorithm and the announced result follows.

\end{document}